\documentclass{article}
\usepackage{spconf,amsmath,graphicx}
\usepackage{subfigure}

\usepackage{enumitem}
\usepackage{wrapfig}
\usepackage{caption}

\title{CONVEX DECOMPOSITION AND EFFICIENT SHAPE REPRESENTATION USING DEFORMABLE CONVEX POLYTOPES}
\name{Fitsum Mesadi and Tolga Tasdizen \thanks{This work is supported by NIH grant 1R01-GM098151-01 and  NSF grant IIS-1149299(TT).}}
\address{Department of Electrical and Computer Engineering, University of Utah, United States}
\begin{document}
%
\maketitle
\begin{abstract}
Decomposition of shapes into (approximate) convex parts is essential for applications such as part-based shape representation, shape matching, and collision detection. In this paper, we propose a novel convex decomposition using a parametric implicit shape model called Disjunctive Normal Shape Model (DNSM). The DNSM is formed as a union of polytopes which themselves are formed by intersections of half-spaces. The key idea is by deforming the polytopes, which naturally remain convex during the evolution, the polytopes capture convex parts without the need to compute convexity. 
 The major contributions of this paper include a robust convex decomposition which also results in an efficient part-based shape representation, and a novel shape convexity measure. The experimental results show the potential of the proposed method. 
\end{abstract}
\section{Introduction}
\label{sec:intro}
Shape segmentation, which is decomposition of shapes into functional and visually meaningful parts, has many applications in areas such as computer vision, graphics, computational geometry and pattern recognition~\cite{Liu2014, Shamir2008, Kaick2014}. One of the most popular geometric constraints in shape segmentation is convexity~\cite{Ren2011, Lien2008}. Decomposition of shapes in to convex components is an active research topic that requires computation of shape convexity measure. Next, we briefly mention some of the related works in the literature on convex decomposition, shape convexity measure, and present the contributions of this paper.

	\noindent
{\bf Convex Decomposition:} 
Decomposition of shapes into exact convex pieces can generate unmanageable number of components due to noise and surface texture~\cite{Ghosh2013}. In this regard, several algorithms have been proposed in the literature for approximate convex decomposition (ACD) with a user specified tolerance for the approximation~\cite{Lien2006, Ren2011, Lien2008, Ghosh2013, Liu2010, Liu2014, Mamou2009}. Lien and Amato in~\cite{Lien2006} presented one of the first significant work in the area of ACD, by recursively resolving the most concave features until the concavity of every component is below some user specified threshold. 
Recently, in~\cite{Liu2010} and~\cite{Ren2013} mutax pairs are used to create fewer and more natural nearly convex shapes. What is common to most of the ACD papers in the literature is that they depend on computation and use of some form of convexity measure for their decomposition.

\noindent
{\bf Convexity Measure:}
Convexity is one of the most basic shape descriptor with many applications~\cite{Lian2012}. The two most common definitions of convexity of a shape are the region-based (RB) and the perimeter-based (PB) approaches. RB defines the convexity of a shape as the ratio of the shape's area to the area of its convex hull; whereas, PB convexity is the ratio of the perimeter of the convex hall of the shape to the perimeter of the shape~\cite{Lian2012, Zunic2004}. However, the major problem of the RB convexity methods are their insensitivity to deep and thin protrusions, while the PB methods are intolerant to small boundary deformations~\cite{Gopalan2010}. Recent papers in the literature that addresses the above problems of the convexity measures of shapes can be found in~\cite{Zunic2004, Lian2012, Gopalan2010, Rahtu2006, Rosin2006}.

\noindent
{\bf Contributions:}
In this paper, we propose a novel convex decomposition and shape convexity measure based on Disjunctive Normal Shape Models (DNSM). The DNSM is an implicit and parametric shape model formed by disjunction of convex polytopes. The convex polytopes are formed by conjunctions of half-spaces (see Fig.~\ref{fig:DNSMIllustration}). 
The DNSM is recently used in different image segmentation approaches~\cite{Mesadi2015, Ramesh2015, Mesadi_ICIP2016, Usman2016}. In the proposed convex decomposition, by deforming the polytopes using variational methods, the polytopes naturally remain convex during the evolution, and hence they capture convex parts without the need to compute convexity. Therefore, unlike most convex decomposition techniques available in the literature that depend on the expensive convexity measure computations on every iteration, the proposed decomposition method generates shape convexity measure as a by-product of its intermediate step. 
This paper has three major contributions: a robust convex decomposition method (Section \ref{sec:METHOD}), a novel shape convexity measure (Section \ref{sec:Convexity}), and an efficient shape representation (Section \ref{sec:EfficientDNSM}). 
By automatically decomposing shapes into their convex parts and representing each part with a convex polytope, we get a compact and geometrically more meaningful DNSM shape representation. 
	\section{Disjunctive Normal Shape Model}
\label{sec:DNSM}
\begin{wrapfigure}{R}{0.12\textwidth}
		\begin{center}
       \subfigure[]{%
            \includegraphics[width=0.07\textwidth]{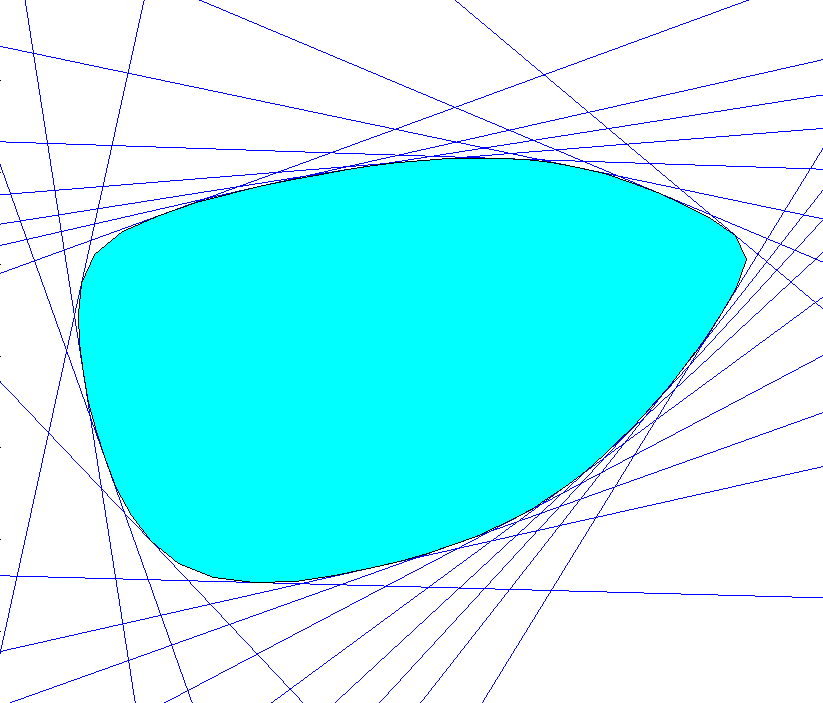}
        }%
					\medskip				
 				\subfigure[]{%
           \includegraphics[width=0.045\textwidth]{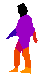}
        }%
   \end{center}
   \vspace{-0.7cm}
    \caption{%
     }%
   \label{fig:DNSMIllustration}
\end{wrapfigure}
In this section, we present the DNSM method which will be used in the next section for the proposed convex decomposition and convexity measures. In DNSM, conjunctions of half spaces form convex polytopes as shown in Fig.~\ref{fig:DNSMIllustration} (a). The disjunction of convex polytopes forms the DNSM shape model as shown in Fig.~\ref{fig:DNSMIllustration} (b) (each color represents a polytope).
  
Consider the characteristic function $f:\mathbf{R}^n\rightarrow\mathbf{B}$ where $\mathbf{B}=\{0,1\}$.  Let $\Omega^+=\{\mathbf{x}\in \mathbf{R}^n:f(\mathbf{x})=1\}$. Let us approximate $\Omega^+$ as the union of $N$ convex polytopes $\tilde{\Omega}^+=\cup_{i=1}^{N} {\cal P}_i$ where the $i^{th}$ polytope is defined as the intersection $P_i=\bigcap_{j=1}^{M}H_{ij}$ of M half-spaces. $H_{ij}$ is defined in terms of its indicator function 
\begin{equation}
h_{ij}(\mathbf{x})=\left\{ 
\begin{array}{lr}
1,& \sum_{k=0}^D w_{ijk}x_k+b_{ij}\geq 0\\
0,& {otherwise}
\end{array}
\right.,
\label{eq:h}
\end{equation}
where $w_{ijk}$ and $b_{ij}$ are the weights and the bias term, and $D$ is the dimension. Therefore, $\Omega^+$ is approximated by $\bigcup_{i=1}^{N}\bigcap_{j=1}^{M}H_{ij}$ and equivalently $f(\mathbf{x})$ is approximated by the disjunctive normal form $\bigvee_{i=1}^{N}\bigwedge_{j=1}^{M} h_{ij}(\mathbf{x})$~\cite{hazewinkel1997}. Converting the disjunctive normal form to a differentiable shape representation requires the following steps: First, De Morgan's rules are used to replace the disjunction with negations and conjunctions, which yields $f(\mathbf{x})\approx \bigvee_{i=1}^{N}\bigwedge_{j=1}^{M} h_{ij}(\mathbf{x})=\neg\bigwedge_{i=1}^N \neg \bigwedge_{j=1}^{M} h_{ij}(\mathbf{x})$. Since conjunctions of binary functions are equivalent to their product and negation is equivalent to subtraction from $1$, $f(\mathbf{x})$ can also be approximated as $1- \prod_{i=1}^{N}(1- \prod_{j=1}^{M}h_{ij}(\mathbf{x}))$. Finally, we approximate $h_{ij}(\mathbf{x})$ with logistic sigmoid functions $\sigma_{ij}(\mathbf{x}) = \frac{1}{1+e^{\sum_{k=0}^D w_{ijk}x_k+b_{ij}}}$ to get the differentiable approximation of the characteristic function $\hat{f}(\mathbf{x})$ 
\vspace{-0.2cm}
\begin{equation}
\hat{f}(\mathbf{x};\mathbf{W}) = 1 - \prod _{i=1}^N 
\left(1-  
\underbrace{
\prod_{j=1}^M 
\frac{1}{1+e^{\sum_{k=0}^D w_{ijk}x_k+b_{ij}}}
}_{g_i(\mathbf{x})}
\right),
\label{eq:DNSM}
\end{equation}
The only adaptive parameters are the weights and biases of the first layer of logistic sigmoid functions, $\sigma_{ij}(\mathbf{x})$, which define the orientations and positions of the linear discriminants that form the shape boundary. 
In equation (\ref{eq:DNSM}) the level set $f (x) = 0.5$ is taken to represent the interface between the foreground $f(\mathbf{x})>0.5$ and background $f(\mathbf{x})<0.5$ regions. 

\section{Convex Decomposition}
\label{sec:METHOD}
The proposed convex decomposition has three steps. First, a given shape is decomposed into many overlapping convex parts starting for a regularly distributed polytopes. Then, we present a novel local convexity measure to sort and remove less significant polytopes. We also show how the local convexity measure can be used as a shape convexity measure. The final convex decomposition is obtained by fitting the DNSM into the shape by using only the selected polytopes. Although we only show for 2D shapes, the proposed algorithm can also be applied to 3D shapes directly.

\subsection{Decomposition into Overlapping Convex Parts}
\label{sec:Decompositionon}
\begin{figure}[ht!] 
		\begin{center}
			\subfigure[]{%
            \includegraphics[width=0.11\textwidth]{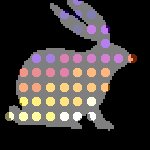}
        }%
				\subfigure[]{%
            \includegraphics[width=0.11\textwidth]{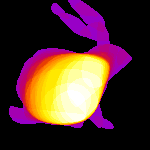}
        }%
 				\subfigure[]{%
           \includegraphics[width=0.11\textwidth]{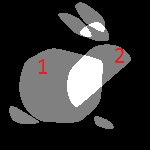}
        }%
				\subfigure[]{%
            \includegraphics[width=0.11\textwidth]{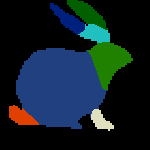}
        }%
   \end{center}
   \vspace{-0.8cm}
    \caption{%
      Demonstration of the proposed convex decomposition steps. a) dense polytopes initialization overlay on the sample shape. b) Result of step 1: Decomposing into overlapping convex parts. c) Result of step 2: Removing polytopes that have small significance measures. d) Result of step 3: Removing the overlaps and filling gaps to get the final decomposition, shown in colors for clarity.
     }%
   \label{fig:DecompositionSteps}
\end{figure}

The goal here is to represent a given shape with overlapping convex polytopes using the DNSM model (\ref{eq:DNSM}). 
We start with large number of polytopes, $N$, as can be seen in Fig.~\ref{fig:DecompositionSteps}(a). The initialization polytopes are approximated as discs (and spheres for 3D) of a fixed radius, and they are regularly distributed in the body of the shape. 
The DNSM discriminant parameters, $\mathbf{W}$, that represent a given shape can be obtained by choosing the weights that minimize the energy
\begin{equation} 
  \begin{split}
 E(\mathbf{W})
&= \int_{\mathbf{x}\in\Omega} \left(f(\mathbf{x})- I(\mathbf{x}\right))^2 d\mathbf{x} \\ 
& - \eta \sum_i\sum_{r \neq i} \int_{\mathbf{x}\in\Omega} g_i(\mathbf{x})g_r(\mathbf{x}) d\mathbf{x} 
	 \label{eq:OverlapDNSMFit}
	  \end{split}
\end{equation}
where $g_i(\mathbf{x})$ represents the individual polytopes of $f(x)$. $I(\mathbf{x})$ is the shape image with intensity value of 1 at the shape and 0 at the background, and $\eta$ is a constant. We minimize (\ref{eq:OverlapDNSMFit}) using gradient descent to obtain $\mathbf{W}$ which represents a given shape, and it will be presented shortly.

The first term in (\ref{eq:OverlapDNSMFit}) fits the model to the shape by minimizing the mean square error between the level set value $f$ and the shape image intensity $I$. The energy from the first term in (\ref{eq:OverlapDNSMFit}) is minimum when $f=1$ inside the object shape (where the intensity value is $I=1$), and $f=0$ outside the object shape (where the intensity value of the ground truth is $I=0$). The second term in (\ref{eq:OverlapDNSMFit}) maximizes the overlap between the different polytopes. 
Fig.~\ref{fig:DecompositionSteps}(b) shows an example of the result obtained by applying equation (\ref{eq:OverlapDNSMFit}) to a shape shown in Fig.~\ref{fig:DecompositionSteps}(a), where the degree of brightness corresponds to the number of polytopes that overlap at that particular point. 
 
The energy minimization implies computing the derivatives of equations of (\ref{eq:OverlapDNSMFit}) with respect to each discriminant parameters, $w_{ijk}$. The update to the discriminant weights, $w_{ijk}$, is then obtained by minimizing the energy using gradient descent as
\begin{equation} 
 \frac{\partial E}{\partial w_{ijk}} =  2 \left(f(\mathbf{x})- I(\mathbf{x})\right) \frac{\partial f}{\partial w_{ijk}} -  \eta \sum_{r \neq i} g_r(\mathbf{x}) \frac{\partial g_i}{\partial w_{ijk}}
\label{eq:dnsm_Region_Der}
\end{equation} 
where $ \frac{\partial f}{\partial w_{ijk}} = \left(1- \prod_{\substack{r \neq i}}(1- g_{r}(\mathbf{x})) \right) \frac{\partial g_{i}}{\partial w_{ijk}} $, and \\
$\frac{\partial g_{i}}{\partial w_{ijk}} = -g_{i}(\mathbf{x}) (1- \sigma_{ij}(\mathbf{x})) x_{k}$, are obtained after a few steps in the taking of the partial derivatives. Therefore, during the evolution the discriminant parameters are updated on each iteration as $w_{ijk} \leftarrow w_{ijk} -\gamma \frac{\partial E}{\partial w_{ijk}} $, where $\gamma$ is the step-size. 

The idea behind maximizing the overlap among the polytopes is that we can easily identify and remove unnecessary polytopes for efficient shape representation and for approximate convex decomposition. For instance, an unnecessary polytope can be a polytope that do not have any unique region it represents. Therefore, removing of such a polytope do not affect the shape representation since the pixels that it represents are already covered by other polytopes. 

\subsection{Shape Convexity Measure}
\label{sec:Convexity}
For approximate convex decomposition, some polytopes are removed based on the ranking of their significance, which is their relative convexity. When exact convex decomposition is needed, the convexity measure to be discussed in this section is not necessary; however, approximate convex decomposition is desirable since it is more robust to (minor) noisy surface deformations, and also results in a compact part-based shape representation. If necessary, the user can control the degree of the approximation by using the number of final polytopes remaining or by the relative convexity measure.

 We define the significance measure of the $i^{th}$ polytope, $C(i)$, as 
\begin{equation} 
  \begin{split}
 C(i) = \frac{R_{Unq}(i)}{R(i)} \left({\frac{R(i)}{R(Largest)}}\right)^{t}
	 \label{eq:LocalConcavity}
	  \end{split}
\end{equation}
\begin{wrapfigure}{R}{0.13\textwidth}
		\begin{center}
       \subfigure[]{%
            \includegraphics[width=0.06\textwidth]{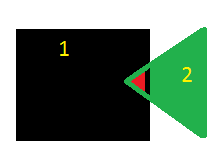}
        }%
					\medskip				
 				\subfigure[]{%
           \includegraphics[width=0.06\textwidth]{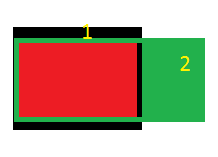}
        }%
   \end{center}
   \vspace{-0.9cm}
    \caption{%
 2 Shapes      }%
   \label{fig:ConvexityDemo}
\end{wrapfigure}
where $R(i)$ is the size of the region represented by the $i^{th}$ polytope, and $R_{Unq}(i)$ is the size of the unique region represented by the $i^{th}$ polytope only. $R(Largest)$ is the size of the largest polytope, and $t$ is a constant which is experimentally found to be around 0.25. For instance, in Fig.~\ref{fig:DecompositionSteps}(c), the polytope labeled $1$ is the largest of all the shown polytopes, and $R(2)$ is the size of polytope 2. The main idea behind equation (\ref{eq:LocalConcavity}) is that the significance of a given polytope depends both on the size of the unique region it represents and on its relative size compared to the largest convex part in the shape.


To show how the significance measure of equation (\ref{eq:LocalConcavity}) is also a local convexity measure, let us look at Fig.~\ref{fig:ConvexityDemo}. The green and black parts are two different polytopes, and the overlap region of the two polytopes is shown in red. In Fig.~\ref{fig:ConvexityDemo}(a) polytope 2 (in green) represents a highly convex local region, and hence it has a large unique region compared to its size, making its $C(2)$ value large. On the other hand, in Fig.~\ref{fig:ConvexityDemo}(b), polytope 2 has a small $C(2)$ value. Therefore, by using deformable polytopes, the true convexity of a local region represented by a given polytope depends on the relative size of the unique region the polytope covers. Note that, both the region-based and the perimeter-based convexity measures defined in Section (\ref{sec:intro}) can not show the large convexity difference between the two shapes in Fig.~\ref{fig:ConvexityDemo}. 
Therefore, based on the local convexity definition of equation (\ref{eq:LocalConcavity}), polytopes that have relatively small $C$ value are removed from further consideration during the approximate convex decomposition. It should be noted that once a given polytope is removed, the unique size of all the remaining polytopes should be recomputed, since the removal of a polytope can affect the unique region sizes of the remaining polytopes. 

Equation (\ref{eq:LocalConcavity}) can be used to measure the (global) concavity of a shape. The global shape concavity is the sum of the local convexity of all the polytopes except the largest polytope; that is, $C_{T}= \sum_{i \neq Largest}C(i)$. Therefore, the proposed global shape concavity depends on the number of convex components the shape has and the relative size of the unique region each polytope represents compared to the largest convex component of the shape. Notice that, the global shape concavity measure is not necessary for the proposed approximate convex decomposition (which only requires local convexity measure); however, convexity of a shape has many applications of its own~\cite{Zunic2004, Lian2012, Gopalan2010}.

\subsection{Final Decomposition and Efficient DNSM}
\label{sec:EfficientDNSM}
The final step in the (approximate) convex decomposition is to represent the shape with DNSM using the few selected polytopes obtained from the significance measure. The shape representation using the DNSM should avoid the overlapping of the polytopes and creation of gaps. For instance, in Fig.~\ref{fig:DecompositionSteps}(c) we can see that removing of many of the 'unnecessary' polytopes has resulted in creation of small gaps, while also some of the polytopes overlap. The energy term that can be minimized, in order to fill the gaps and remove the overlaps using the final small number of selected polytopes, can be given as   
\begin{equation} 
  \begin{split}
 E(\mathbf{W})
&= \int_{\mathbf{x}\in\Omega} \left(f(\mathbf{x})- I(\mathbf{x}\right))^2 d\mathbf{x} \\ 
& + \eta \sum_i\sum_{r \neq i} \int_{\mathbf{x}\in\Omega} g_i(\mathbf{x})g_r(\mathbf{x}) d\mathbf{x} 
	 \label{eq:FinalDNSMFit}
	  \end{split}
\end{equation}
which is similar to equation (\ref{eq:OverlapDNSMFit}) except for the plus sign in front of the second term. That is, in equation (\ref{eq:FinalDNSMFit}) we penalize the overlap of the polytopes in order to represent each part with a unique convex polytope. The first term in equation (\ref{eq:FinalDNSMFit}) helps to fill the gaps and fits the DNSM model to the shape as discussed previously for (\ref{eq:OverlapDNSMFit}). Figure~\ref{fig:DecompositionSteps}(d) shows the  final convex decomposed result obtained by applying equation (\ref{eq:FinalDNSMFit}) to the result in Fig.~\ref{fig:DecompositionSteps}(c). Therefore, by decomposing and representing each convex part of a given shape with a unique polytope we achieve a very compact and geometrically meaningful shape representation. For instance, in Fig.~\ref{fig:DecompositionSteps}(d) only $N=7$ polytopes (each of which has $M=16$ discriminants) are needed to represent the shape with great accuracy, resulting in large compression rate. In addition, by storing the connectivity graph of the polytopes (that is, which poltope is connected to which), we can construct a graphical DNSM shape representation that can facilitate further shape analysis algorithms such as shape matching and recognition.

\section{RESULTS}
\label{sec:RESULTS}
In this section, we present the experimental results for the proposed convex decomposition and the shape convexity measure. 

\begin{figure}[ht!] 
\captionsetup[subfigure]{labelformat=empty}
		\begin{center}

				\subfigure{%
            \includegraphics[width=0.09\textwidth]{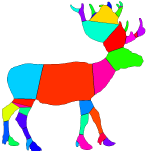}
        }%
				\subfigure{%
            \includegraphics[width=0.09\textwidth]{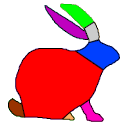}
        }%
				\subfigure{%
            \includegraphics[width=0.09\textwidth]{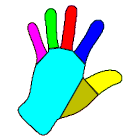}
        }%
 				\subfigure{%
           \includegraphics[width=0.09\textwidth]{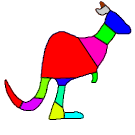}
        }%
				\subfigure{%
            \includegraphics[width=0.09\textwidth]{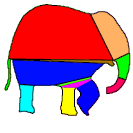}
        }%
				
				   \vspace{-0.4cm}					
								\medskip	
				\subfigure{%
            \includegraphics[width=0.09\textwidth]{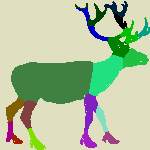}
        }
				\subfigure{%
            \includegraphics[width=0.09\textwidth]{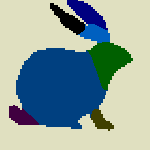}
        }%
				\subfigure{%
            \includegraphics[width=0.09\textwidth]{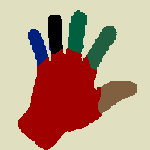}
        }%
 				\subfigure{%
           \includegraphics[width=0.09\textwidth]{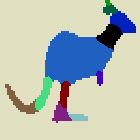}
        }%
				\subfigure{%
            \includegraphics[width=0.09\textwidth]{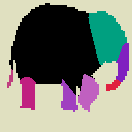}
        }%
   \end{center}
   \vspace{-0.8cm}
    \caption{%
      First row (top) are the decomposition results using~\cite{Lien2006}, and in the second row are the results using our DNSM.
     }%
   \label{fig:Decomposition}
\end{figure}

\begin{figure}[ht!] 
\captionsetup[subfigure]{labelformat=empty}
		\begin{center}
%
				\subfigure{%
            \includegraphics[width=0.09\textwidth]{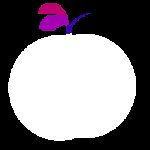}
        }%
 				\subfigure{%
           \includegraphics[width=0.09\textwidth]{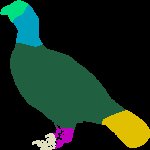}
        }%
				\subfigure{%
            \includegraphics[width=0.09\textwidth]{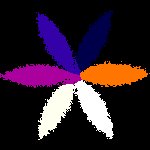}
        }%
				\subfigure{%
            \includegraphics[width=0.09\textwidth]{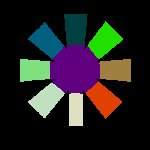}
        }%
				 \subfigure{%
           \includegraphics[width=0.09\textwidth]{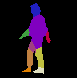}
        }%
   \end{center}
   \vspace{-0.7cm}
    \caption{%
            Results using the proposed DNSM decomposition     }%
   \label{fig:FurtherDecompositionExamples}
\end{figure}

\noindent
{\bf Decomposition Results:}  
Figure~\ref{fig:Decomposition} shows the shape decomposition examples using the proposed method, and compares them with the result using Lien et al.~\cite{Lien2006}. As can be seen from the figure, the results using the proposed method (Fig.~\ref{fig:Decomposition} second row) shows convex decomposition that is closer to human expectation compared to results using~\cite{Lien2006}. Our algorithm also decomposes the shapes in to smaller number of parts compared to~\cite{Lien2006}, which is essential for robust shape representation and can improve the efficiency of further processes. 
Figure~\ref{fig:FurtherDecompositionExamples} gives additional convex decomposition examples for shapes from MPEG-7 dataset~\cite{Latecki2000} and a walking person, using the proposed method. 

\noindent
{\bf Convexity Results:}  
We compare the concavity measure proposed in this paper with the two commonly used concavity measures in the literature: the PB and the RB concavities as defined in Section (\ref{sec:intro}). Figure~\ref{fig:Decomposition} gives a comparison of the three concavity measures using shapes from MPEG-7 dataset~\cite{Latecki2000}. Looking at the shapes in the figure, one can easy see that the apple is the least concave (highly convex) followed by the birds, and then the camels, and finally the star device. However, the PB and RB concavity measures made mistakes both in ranking the different object classes and the positions of objects of the same class (for instance, the two birds) that have very small concavity difference when observed by humans. The proposed DNSM-based concavity measure and its ranking corresponds more closely to what is expected.
\vspace{-0.4cm}
\begin{figure}[ht!] 
\captionsetup[subfigure]{labelformat=empty}
		\begin{center}
				\subfigure{%
            \includegraphics[width=0.07\textwidth]{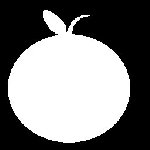}
        }%
				\subfigure{%
            \includegraphics[width=0.07\textwidth]{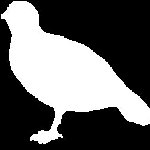}
        }%
				\subfigure{%
            \includegraphics[width=0.07\textwidth]{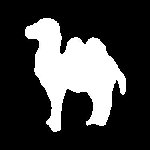}
        }%
				\subfigure{%
           \includegraphics[width=0.07\textwidth]{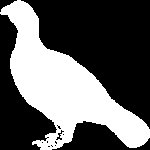}
        }%
				\subfigure{%
            \includegraphics[width=0.07\textwidth]{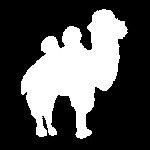}
        }%
				\subfigure{%
           \includegraphics[width=0.07\textwidth]{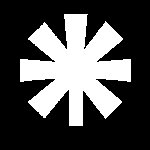}
        }%
								\medskip	

				\subfigure{%
            \includegraphics[width=0.07\textwidth]{apple-7.jpg}
        }%
 				\subfigure{%
           \includegraphics[width=0.07\textwidth]{camel-1.jpg}
        }%
 				\subfigure{%
           \includegraphics[width=0.07\textwidth]{camel-12.jpg}
        }%
				\subfigure{%
            \includegraphics[width=0.07\textwidth]{bird-2.jpg}
        }%
				\subfigure{%
           \includegraphics[width=0.07\textwidth]{Sdevice2-18.jpg}
        }%
					\subfigure{%
           \includegraphics[width=0.07\textwidth]{Sbird-7.jpg}
        }%
							\medskip	

				\subfigure{%
            \includegraphics[width=0.07\textwidth]{apple-7.jpg}
        }%
 				\subfigure{%
           \includegraphics[width=0.07\textwidth]{bird-2.jpg}
        }%
				\subfigure{%
           \includegraphics[width=0.07\textwidth]{Sbird-7.jpg}
        }%
								\subfigure{%
            \includegraphics[width=0.07\textwidth]{camel-12.jpg}
        }%
 				\subfigure{%
           \includegraphics[width=0.07\textwidth]{camel-1.jpg}
        }%
					\subfigure{%
           \includegraphics[width=0.07\textwidth]{Sdevice2-18.jpg}
        }%
   \end{center}
   \vspace{-0.8cm}
    \caption{%
      Comparison of concavity measures. The shapes are ranked in increasing concavity from left to right. Top row is ranking using perimeter-based Concavity (PBC), middle row using region-based concavity (RBC), and bottom row is using our DNSM concavity measure. The concavity values in numbers are as follows. PBC left to right 0.1488, 0.1707, 0.3077, 0.3180, 0.3819, 0.5265. RBC left to right 0.1023, 0.2482, 0.2891, 0.2901, 0.3017, 0.3023. DNSM left to right 0.5095, 1.0234, 2.1478, 4.2576, 5.8605, 8.6839.	
     }%
   \label{fig:ConcavityComparison}
 \vspace{-0.4cm}
\end{figure}   
 
\section{Conclusion}
\label{sec:Conclusion}
In this paper, we presented a novel convex decomposition by using deformable convex polytopes, which naturally maintain their convexity during deformation. 
This is conceptually different from the techniques commonly available in the literature which require the computation of convexity on every iteration, which can be expensive and may not be reliable. The proposed decomposition method generates a shape convexity measure, which corresponds well to the convexity observed by humans,  as a by-product of its intermediate step. The approximate convex decomposition helps to form a compact and efficient DNSM, where each convex part is represented by a single polytope. In the future, we plan to extend the method presented in this paper to shape matching and recognition.

\bibliographystyle{IEEEbib}
\bibliography{strings,refsICIP}

\begin{thebibliography}{10}

\bibitem{Liu2014}
Guilin Liu, Zhonghua Xi, and Jyh-Ming Lien,
\newblock ``Dual-space decomposition of 2d complex shapes,''
\newblock in {\em Computer Vision and Pattern Recognition (CVPR), 2014 IEEE
  Conference on}, June 2014, pp. 4154--4161.

\bibitem{Shamir2008}
Ariel Shamir,
\newblock ``A survey on mesh segmentation techniques,''
\newblock {\em Computer Graphics Forum}, vol. 27, no. 6, pp. 1539--1556, 2008.

\bibitem{Kaick2014}
Oliver~Van Kaick, Noa Fish, Yanir Kleiman, Shmuel Asafi, and Daniel Cohen-OR,
\newblock ``Shape segmentation by approximate convexity analysis,''
\newblock {\em ACM Trans. Graph.}, vol. 34, no. 1, pp. 4:1--4:11, Dec. 2014.

\bibitem{Ren2011}
Zhou Ren, Junsong Yuan, Chunyuan Li, and Wenyu Liu,
\newblock ``Minimum near-convex decomposition for robust shape
  representation,''
\newblock in {\em Computer Vision (ICCV), 2011 IEEE International Conference
  on}, Nov 2011, pp. 303--310.

\bibitem{Lien2008}
Jyh-Ming Lien and Nancy~M. Amato,
\newblock ``Approximate convex decomposition of polyhedra and its
  applications,''
\newblock {\em Computer Aided Geometric Design}, vol. 25, no. 7, pp. 503 --
  522, 2008,
\newblock Solid and Physical ModelingSelected papers from the Solid and
  Physical Modeling and Applications Symposium 2007 (SPM 2007)Solid and
  Physical Modeling and Applications Symposium 2007.

\bibitem{Ghosh2013}
Mukulika Ghosh, Nancy~M. Amato, Yanyan Lu, and Jyh-Ming Lien,
\newblock ``Fast approximate convex decomposition using relative concavity,''
\newblock {\em Computer-Aided Design}, vol. 45, no. 2, pp. 494 -- 504, 2013,
\newblock Solid and Physical Modeling 2012.

\bibitem{Lien2006}
Jyh-Ming Lien and Nancy~M. Amato,
\newblock ``Approximate convex decomposition of polygons,''
\newblock {\em Computational Geometry}, vol. 35, no. 1–2, pp. 100 -- 123,
  2006,
\newblock Special Issue on the 20th \{ACM\} Symposium on Computational
  Geometry20th \{ACM\} Symposium on Computational Geometry.

\bibitem{Liu2010}
Hairong Liu, Wenyu Liu, and L.J. Latecki,
\newblock ``Convex shape decomposition,''
\newblock in {\em Computer Vision and Pattern Recognition (CVPR), 2010 IEEE
  Conference on}, June 2010, pp. 97--104.

\bibitem{Mamou2009}
K.~Mamou and F.~Ghorbel,
\newblock ``A simple and efficient approach for 3d mesh approximate convex
  decomposition,''
\newblock in {\em Image Processing (ICIP), 2009 16th IEEE International
  Conference on}, Nov 2009, pp. 3501--3504.

\bibitem{Ren2013}
Zhou Ren, Junsong Yuan, and Wenyu Liu,
\newblock ``Minimum near-convex shape decomposition,''
\newblock {\em IEEE Trans. on on Pattern Analysis and Machine Intelligence},
  vol. 35, pp. 2546--2552, 2013.

\bibitem{Lian2012}
Zhouhui Lian, A.~Godil, P.L. Rosin, and Xianfang Sun,
\newblock ``A new convexity measurement for 3d meshes,''
\newblock in {\em Computer Vision and Pattern Recognition (CVPR), 2012 IEEE
  Conference on}, June 2012, pp. 119--126.

\bibitem{Zunic2004}
J.~Zunic and P.L. Rosin,
\newblock ``A new convexity measure for polygons,''
\newblock {\em Pattern Analysis and Machine Intelligence, IEEE Transactions
  on}, vol. 26, no. 7, pp. 923--934, July 2004.

\bibitem{Gopalan2010}
Raghuraman Gopalan, Pavan Turaga, and Rama Chellappa,
\newblock ``Articulation-invariant representation of non-planar shapes,''
\newblock in {\em Proceedings of the 11th European Conference on Computer
  Vision Conference on Computer Vision: Part III}, Berlin, Heidelberg, 2010,
  ECCV'10, pp. 286--299, Springer-Verlag.

\bibitem{Rahtu2006}
E.~Rahtu, M.~Salo, and J.~Heikkila,
\newblock ``A new convexity measure based on a probabilistic interpretation of
  images,''
\newblock {\em Pattern Analysis and Machine Intelligence, IEEE Transactions
  on}, vol. 28, no. 9, pp. 1501--1512, Sept 2006.

\bibitem{Rosin2006}
Paul~L. Rosin and Christine~L. Mumford,
\newblock ``A symmetric convexity measure,''
\newblock {\em Computer Vision and Image Understanding}, vol. 103, no. 2, pp.
  101 -- 111, 2006.

\bibitem{Mesadi2015}
F.~Mesadi, M.~Cetin, and T.~Tasdizen,
\newblock ``Disjunctive normal shape and appearance priors with applications to
  image segmentation,''
\newblock in {\em Medical Image Computing and Computer-Assisted Intervention
  — MICCAI 2015}, Nassir Navab, Joachim Hornegger, WilliamM. Wells, and
  AlejandroF. Frangi, Eds., vol. 9351 of {\em Lecture Notes in Computer
  Science}, pp. 703--710. Springer International Publishing, 2015.

\bibitem{Ramesh2015}
M.~Ramesh, F.~Mesadi, M.~Cetin, and T.~Tasdizen,
\newblock ``Disjunctive normal shape model,''
\newblock in {\em ISBI, IEEE International Symposium on Biomedical Imaging},
  2015.

\bibitem{Mesadi_ICIP2016}
F.~Mesadi and T.~Tasdizen,
\newblock ``Disjunctive normal level set: An efficient parametric implicit
  method,''
\newblock in {\em 2016 IEEE International Conference on Image Processing
  (ICIP)}, Oct 2016.

\bibitem{Usman2016}
M.~Ghani, F.~Mesadi, S.~Kanık, Argunsah A., Israely I., M.~Cetin, and
  T.~Tasdizen,
\newblock ``Dendritic spine shape analysis using disjunctive normal shape
  model,''
\newblock in {\em ISBI, IEEE International Symposium on Biomedical Imaging},
  2016.

\bibitem{hazewinkel1997}
M.~Hazewinkel,
\newblock {\em Encyclopaedia of Mathematics: An Updated and Annotated
  Translation of the Soviet "Mathematical Encyclopaedia},
\newblock Number v. 1 in Encyclopaedia of Mathematics. Springer, 1997.

\bibitem{Latecki2000}
L.J. Latecki, R.~Lakamper, and T.~Eckhardt,
\newblock ``Shape descriptors for non-rigid shapes with a single closed
  contour,''
\newblock in {\em Computer Vision and Pattern Recognition, 2000. Proceedings.
  IEEE Conference on}, 2000, vol.~1, pp. 424--429 vol.1.

\end{thebibliography}

\end{document}